\documentclass[11pt]{article}

\usepackage[preprint]{acl}

\usepackage{times}
\usepackage{latexsym}

\usepackage[T1]{fontenc}

\usepackage[utf8]{inputenc}

\usepackage{microtype}

\usepackage{inconsolata}

\usepackage{graphicx}

\usepackage{enumitem}
\usepackage{algorithm}
\usepackage{algorithmic}
\usepackage{booktabs}
\usepackage{amsfonts}
\usepackage{amsmath}
\usepackage{float}
\usepackage{multirow}
\usepackage{subcaption}

\newcommand\blfootnote[1]{%
  \begingroup
  \renewcommand\thefootnote{}
  \footnotetext{#1}
  \endgroup
}

\title{Reasoning Stabilization Point: A Training-Time Signal for Stable Evidence and Shortcut Reliance}

\author{Sahil Rajesh Dhayalkar \\
  Arizona State University \\
  \texttt{sdhayalk@asu.edu} \\}

\begin{document}
\maketitle
\begin{abstract}
Fine-tuning pretrained language models can improve task performance while subtly altering the evidence a model relies on. We propose a training-time interpretability view that tracks token-level attributions across fine-tuning epochs. We define \emph{explanation drift} as the epoch-to-epoch change in normalized token attributions on a fixed probe set, and introduce the \emph{Reasoning Stabilization Point} (RSP), the earliest epoch after which drift remains consistently low. RSP is computed from within-run drift dynamics and requires no tuning on out-of-distribution data. Across multiple lightweight transformer classifiers and benchmark classification tasks, drift typically collapses into a low, stable regime early in training, while validation accuracy continues to change only marginally. In a controlled shortcut setting with label-correlated trigger tokens, attribution dynamics expose increasing reliance on the shortcut even when validation accuracy remains competitive. Overall, explanation drift provides a simple, low-cost diagnostic for monitoring how decision evidence evolves during fine-tuning and for selecting checkpoints in a stable-evidence regime.
\end{abstract}

\section{Introduction}
\blfootnote{This paper is under review at ACL Rolling Review}Fine-tuning pretrained language models delivers strong performance with modest compute, but it can also change \emph{how} a model arrives at decisions in ways that are not visible from accuracy alone. In NLP, models often exploit dataset artifacts and shallow heuristics that perform well in-domain yet fail under distribution shift~\cite{gururangan-etal-2018-annotation,mccoy-etal-2019-right,geirhos-etal-2020-shortcut,poliak-etal-2018-hypothesis,glockner-etal-2018-breaking,niven-kao-2019-probing}. This motivates interpretability methods that expose decision evidence, such as feature-attribution and saliency~\cite{ribeiro-etal-2016-trust,sundararajan-etal-2017-axiomatic,lundberg-lee-2017-unified,NIPS2014_ca007296}. However, explanations are typically reported as \emph{static snapshots} (e.g., after selecting the best checkpoint), leaving open a basic question: \textbf{how do explanations evolve during fine-tuning time?}

We propose a lightweight, training-time interpretability view: treat token-level explanations as a time series over epochs. Intuitively, if fine-tuning is progressively ``locking in'' decision evidence, attributions should become stable; conversely, continued changes in evidence with small accuracy gains may reflect shifting reliance toward brittle cues. Concretely, we define \emph{explanation drift} as the change in normalized token attributions between consecutive epochs on a fixed probe set, and we summarize the drift curve by a single interpretable event called the \emph{Reasoning Stabilization Point} (RSP), the earliest epoch after which drift remains low for a short window. RSP is fully determined by the drift dynamics of a run and does not require tuning on OOD (Out-Of-Distribution) data.

Our focused contributions are:
\begin{itemize}[itemsep=0pt, topsep=0pt]
    \item \textbf{Explanation Drift:} a model-agnostic scalar curve measuring epoch-to-epoch change in token attributions on a frozen probe set.
    \item \textbf{Reasoning Stabilization Point (RSP):} a compact training-time signal that marks when token-importance patterns become stable.
    \item \textbf{Empirical evidence:} drift stabilizes by (or before) accuracy saturation and reveals shortcut reliance hidden by validation accuracy.
\end{itemize}

Because the method is post-hoc over saved checkpoints and uses off-the-shelf gradient-based attributions, it adds minimal overhead beyond standard fine-tuning. We view RSP as a practical diagnostic for deciding \emph{when} a model's decision evidence stabilizes, and for surfacing shortcut adoption that may be hidden by aggregate accuracy.

\section{Related Work}
\paragraph{Post-hoc explanations in NLP.}
A large body of work studies explanations via feature attribution ~\cite{ancona2018towards,li-etal-2016-visualizing,arras-etal-2017-explaining} and local surrogate models, including LIME~\cite{ribeiro-etal-2016-trust} and Shapley-based frameworks such as SHAP~\cite{lundberg-lee-2017-unified}. Gradient-based attributions (e.g., Integrated Gradients) are widely used for neural models because they are simple and require no auxiliary explainer~\cite{sundararajan-etal-2017-axiomatic}. Parallel to attribution, rationale-based approaches aim to produce human-readable evidence, with benchmarks such as ERASER supporting systematic evaluation of rationales and faithfulness~\cite{deyoung-etal-2020-eraser}. Our work is complementary: we do not propose a new explainer or rationale extractor, but rather a \emph{temporal} lens over any token-level explanation function.

\paragraph{Faithfulness, sanity checks, and stability of explanations.}
The reliability of explanations has been scrutinized via sanity checks and robustness criteria.~\cite{adebayo-etal-2018-sanity} showed that some saliency maps can be insensitive to model parameters or data, motivating careful evaluation of explanation methods. Robustness of interpretability methods, whether similar inputs yield similar explanations, has also been formalized and empirically tested~\cite{alvarez-melis-jaakkola-2018-robustness}. Debate around using internal signals such as attention as explanations further underscores the need for principled comparison and diagnostics~\cite{jain-wallace-2019-attention,wiegreffe-pinter-2019-attention,serrano-smith-2019-attention}. Our approach differs from input-perturbation robustness: we measure \emph{stability across training time} on a fixed probe set, asking when the model's token-importance profile becomes consistent across epochs.

\paragraph{Shortcut learning, dataset artifacts, and training dynamics.}
Models can achieve high benchmark performance by exploiting spurious correlations and annotation artifacts~\cite{gururangan-etal-2018-annotation}, and NLI systems in particular often adopt shallow heuristics that fail on challenge sets such as HANS~\cite{mccoy-etal-2019-right}. More broadly, shortcut learning has been argued to underlie many failures of generalization~\cite{geirhos-etal-2020-shortcut}. Prior tools for understanding training behavior include data-centric analyses (e.g., influence functions linking predictions to training points)~\cite{koh-liang-2017-influence}. Our work contributes a complementary, low-cost \emph{training-time} diagnostic: explanation drift and RSP summarize how decision evidence shifts (or stabilizes) as fine-tuning proceeds, and can help identify when continued fine-tuning may be reallocating importance toward brittle cues rather than learning robust features.

\section{Explanation Drift Across Fine-Tuning Time}
\label{sec:method}

\paragraph{Main claim.}
We study \emph{how} a fine-tuned classifier arrives at its predictions over training time, not just \emph{whether} it is accurate. Our central claim is that token-level explanations (attributions) form a meaningful time series during fine-tuning and often \emph{stabilize earlier} than peak in-domain validation accuracy. We operationalize this phenomenon via \emph{explanation drift}, the change in attribution patterns between successive epochs, and define a single interpretable training-time event, the \emph{Reasoning Stabilization Point (RSP)}, as the first epoch where drift becomes small and remains small. Intuitively, RSP marks when the model ``locks in'' a relatively stable set of decision features. When fine-tuning continues past RSP, we often observe diminishing accuracy gains alongside non-trivial residual drift, consistent with the model reallocating importance toward brittle or shortcut features. A practical corollary is that early stopping at (or near) RSP can preserve or improve robustness (e.g., OOD/challenge performance) while matching in-domain accuracy (TODO can perhaps remove this corollary)

\subsection{Setup and token-level explanations}
Let $f_{\theta_t}$ denote a classifier after epoch $t\in\{1,\ldots,T\}$ of fine-tuning, where $\theta_t$ are the model parameters at epoch $t$. We fix a \emph{probe set} $P=\{(x_i,y_i)\}_{i=1}^n$ (e.g., a small held-out subset of the development set) that remains constant across epochs. Fixing $P$ ensures that changes in explanations reflect changes in the model, not changes in sampled inputs.

For an input sequence $x=(w_1,\ldots,w_{|x|})$, we compute a token-level explanation
\begin{equation}
\begin{aligned}
& E_{\theta}(x) \in \mathbb{R}^{|x|}, \\
& E_{\theta}(x)_j \;\; \text{is the attribution for token } w_j.
\end{aligned}
\end{equation}
We instantiate $E_{\theta}$ using standard gradient-based attribution methods over the embedding layer (e.g., integrated gradients or gradient$\times$input), which are lightweight and require no auxiliary models. To compare explanations across epochs, we convert raw attributions into a normalized distribution over tokens. Specifically, we take absolute values (capturing magnitude of influence regardless of sign) and normalize onto the simplex:
\begin{equation}
\tilde{E}_{\theta}(x)_j \;=\; \frac{\left|E_{\theta}(x)_j\right|}{\sum_{k=1}^{|x|}\left|E_{\theta}(x)_k\right| + \varepsilon},
\label{eq:normalize}
\end{equation}
where $\varepsilon$ is a small constant for numerical stability. This makes $\tilde{E}_{\theta}(x)$ comparable across epochs and across inputs of different lengths.

\subsection{Explanation drift}
We quantify how a model's \emph{reasoning evidence} changes from epoch to epoch by measuring the discrepancy between successive normalized explanations on the same input. For each example $x\in P$, we define the \emph{per-example drift} between epochs $t-1$ and $t$ as
\begin{equation}
d_t(x) \;=\; 1 - \rho\!\left(\tilde{E}_{\theta_t}(x), \tilde{E}_{\theta_{t-1}}(x)\right),
\label{eq:per_example_drift}
\end{equation}
where $\rho(\cdot,\cdot)$ is a similarity measure between two attribution vectors. In our implementation, we use Spearman rank correlation, which is robust to monotone transformations and focuses on whether the \emph{ordering} of important tokens changes. (Cosine similarity is a drop-in alternative.)

We then aggregate drift over the probe set to obtain a single scalar per epoch:
\begin{equation}
D_t \;=\; \frac{1}{|P|}\sum_{x\in P} d_t(x).
\label{eq:dataset_drift}
\end{equation}
$D_t$ is an \emph{explanation drift curve} over training time. High $D_t$ indicates that the model is still reallocating importance across tokens, while low $D_t$ indicates stable attribution patterns. For interpretability, one can optionally compute label-conditional drift $D_t^{(y)}$ by averaging over $\{x\in P: y(x)=y\}$, revealing whether stabilization occurs uniformly across classes.

\subsection{Reasoning Stabilization Point (RSP)}
We convert the drift curve into a training-time event. Let $w$ be a small window size (we use $w\in\{2,3\}$). We define the \emph{Reasoning Stabilization Point (RSP)} as the earliest epoch after which drift remains below a threshold for $w$ consecutive epochs:
\begin{equation}
\mathrm{RSP} \;=\; \min\left\{ t \;:\; \frac{1}{w}\sum_{k=0}^{w-1} D_{t+k} \le \tau \right\}.
\label{eq:rsp}
\end{equation}
The threshold $\tau$ is chosen \emph{data-driven} to avoid tuning on OOD or challenge sets. A simple and effective rule is
\begin{equation}
\tau \;=\; \mathrm{median}(D_2,\ldots,D_T),
\end{equation}
which anchors stabilization to the typical drift observed over the fine-tuning run. RSP is therefore fully determined by the drift dynamics of the run and a small fixed window size.

RSP is the first epoch where the model's token-importance profile becomes consistently stable on the probe set. This operationalizes the notion that a model may reach a point where its ``reasons'' for predictions no longer change much, even if accuracy continues to improve marginally. In later epochs, non-trivial drift with small accuracy gains can indicate that the model is reallocating attention toward fragile cues, providing an interpretable signal for early stopping and shortcut detection.



\begin{algorithm}[t]
\caption{Explanation drift and RSP}
\label{alg:rsp}
\begin{algorithmic}[1]
\REQUIRE Fine-tuning checkpoints $\{\theta_t\}_{t=1}^T$, probe set $P$, window $w$, threshold rule for $\tau$
\FOR{$t=1$ to $T$}
    \FOR{each $x\in P$}
        \STATE Compute token attributions $E_{\theta_t}(x)$
        \STATE Normalize to $\tilde{E}_{\theta_t}(x)$ via Eq.~\ref{eq:normalize}
    \ENDFOR
\ENDFOR
\FOR{$t=2$ to $T$}
    \STATE Compute $D_t$ using Eq.~\ref{eq:per_example_drift}--\ref{eq:dataset_drift}
\ENDFOR
\STATE Set $\tau$ (e.g., $\tau=\mathrm{median}(D_2,\ldots,D_T)$)
\STATE Compute $\mathrm{RSP}$ using Eq.~\ref{eq:rsp}
\RETURN Drift curve $\{D_t\}_{t=2}^T$, $\mathrm{RSP}$, (optional) 
\end{algorithmic}
\end{algorithm}

\section{Experiments}
\label{sec:experiments}

\subsection{Experimental setup}
\label{subsec:experimental_setup}
We evaluate whether explanation drift stabilizes before (or at least by) accuracy saturation, and whether drift surfaces shortcut adoption. We fine-tune each of the two models~\textsc{DistilBERT}~\cite{distilbert}(variant ~\texttt{distilbert-base-uncased}) and~\textsc{MiniLM} (variant \texttt{microsoft/MiniLM-L12-H384-uncased}) ~\cite{minilm} on SST-2~\cite{sst2} and QNLI~\cite{qnli} (GLUE~\cite{glue}) for $T{=}5$ epochs. At each epoch $t$, we compute token attributions using gradient$\times$input at the embedding layer for the gold label, normalize them to $\tilde{E}_{\theta_t}(x)$ (Eq.~2), and measure drift $D_t$ via Spearman similarity (Eq.~3--4). We fix a probe set of $n{=}500$ validation examples across epochs and seeds. RSP uses a short window $w{=}2$ and the data-driven threshold $\tau=\mathrm{median}(D_2,\ldots,D_T)$ (Eq.~5--6). We repeat all experiments over $3$ random seeds and report mean$\pm$std. Hyperparameters include using a batch size of $128$, tokenizer length $128$, AdamW as optimizer with learning rate of $2 \times 10^{-5}$ and weight decay of $0.01$. All experiments are implemented in PyTorch~\cite{pytorch} and are executed using CUDA acceleration on an NVIDIA GeForce RTX 4060 GPU.

\subsection{Experiment 1: Does explanation stabilize before accuracy?}
\label{subsec:exp1}
\begin{figure*}[t] 
    \centering
    
    \begin{subfigure}{0.32\textwidth}
        \includegraphics[width=\linewidth]{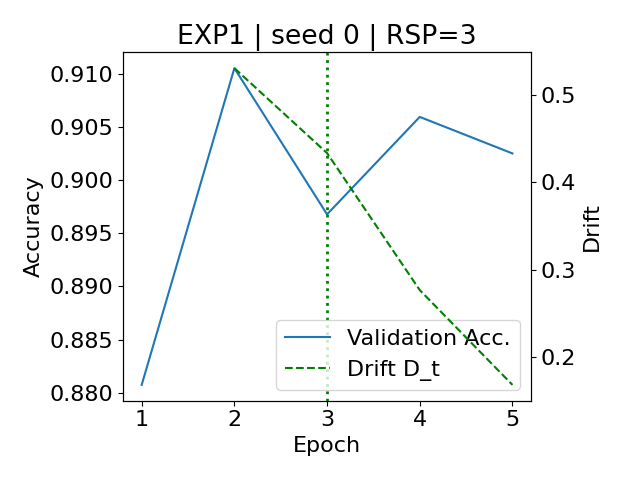}
        \caption{~\textsc{DistilBERT} + SST-2; seed=0}
    \end{subfigure}
    \hfill
    \begin{subfigure}{0.32\textwidth}
        \includegraphics[width=\linewidth]{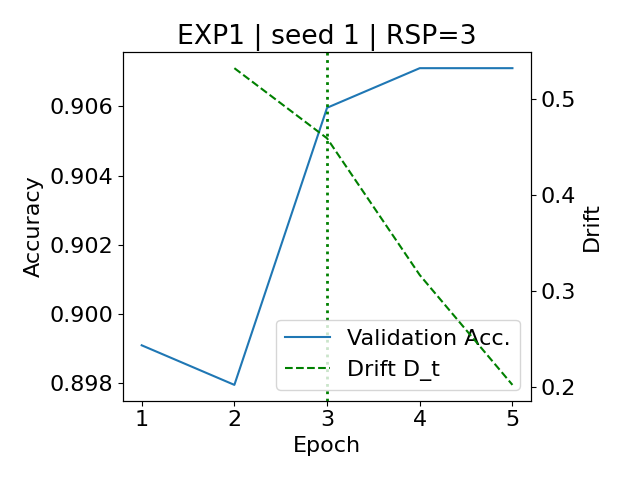}
        \caption{~\textsc{DistilBERT} + SST-2; seed=1}
    \end{subfigure}
    \hfill
    \begin{subfigure}{0.32\textwidth}
        \includegraphics[width=\linewidth]{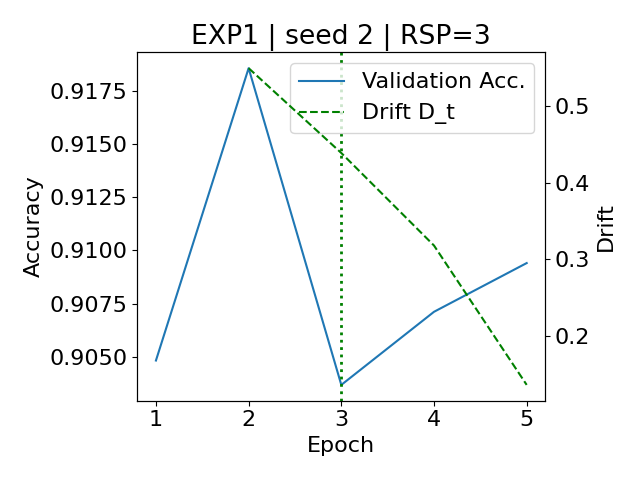}
        \caption{~\textsc{DistilBERT} + SST-2; seed=2}
    \end{subfigure}

    \vspace{1em} 

    \begin{subfigure}{0.32\textwidth}
        \includegraphics[width=\linewidth]{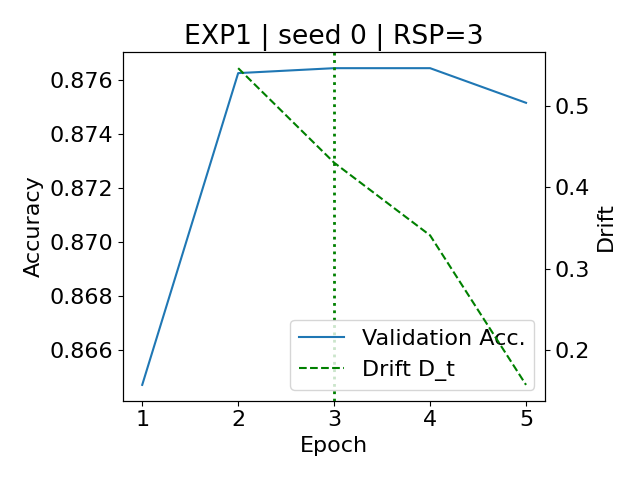}
        \caption{~\textsc{DistilBERT} + QNLI; seed=0}
    \end{subfigure}
    \hfill
    \begin{subfigure}{0.32\textwidth}
        \includegraphics[width=\linewidth]{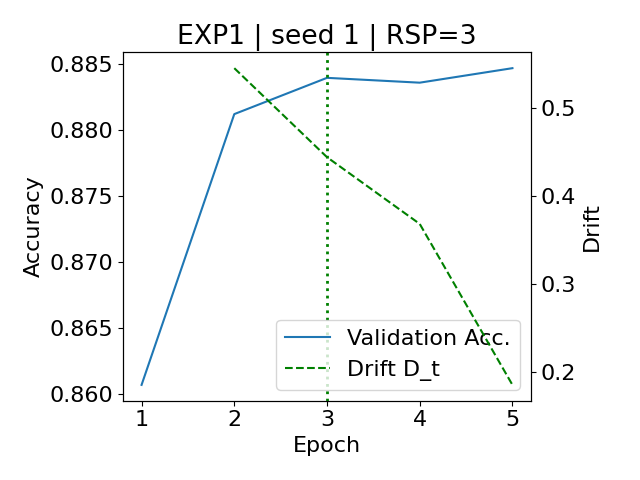}
        \caption{~\textsc{DistilBERT} + QNLI; seed=1}
    \end{subfigure}
    \hfill
    \begin{subfigure}{0.32\textwidth}
        \includegraphics[width=\linewidth]{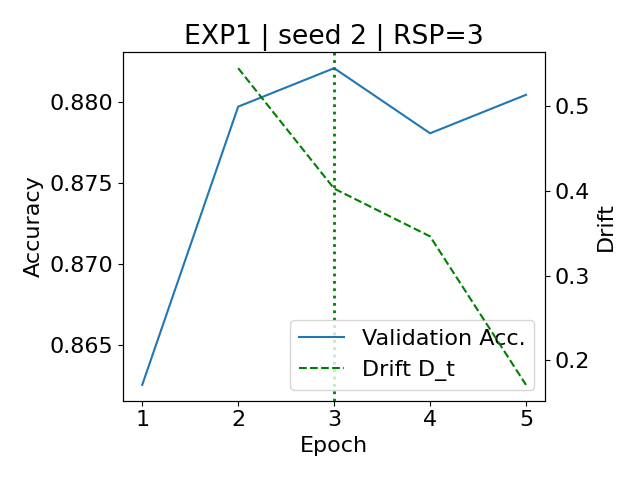}
        \caption{~\textsc{DistilBERT} + QNLI; seed=2}
    \end{subfigure}

    \vspace{1em}

    \begin{subfigure}{0.32\textwidth}
        \includegraphics[width=\linewidth]{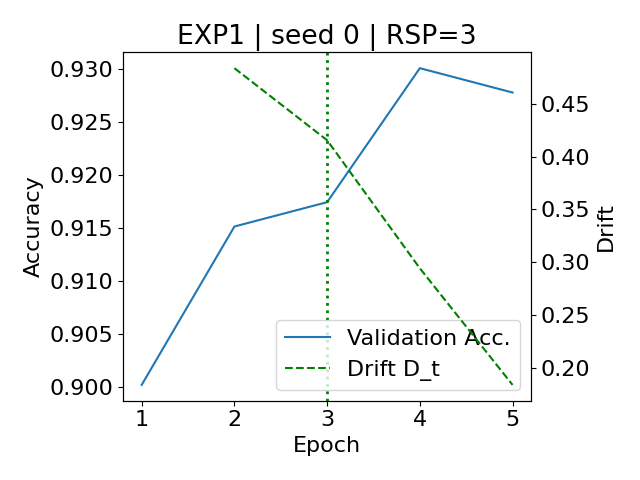}
        \caption{~\textsc{MiniLM} + SST-2; seed=0}
    \end{subfigure}
    \hfill
    \begin{subfigure}{0.32\textwidth}
        \includegraphics[width=\linewidth]{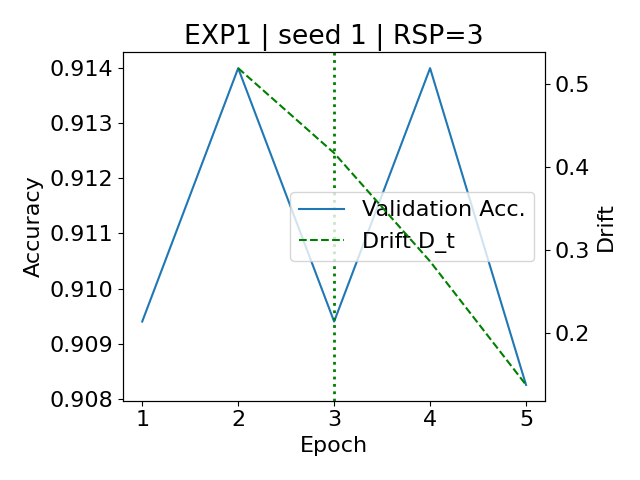}
        \caption{~\textsc{MiniLM} + SST-2; seed=1}
    \end{subfigure}
    \hfill
    \begin{subfigure}{0.32\textwidth}
        \includegraphics[width=\linewidth]{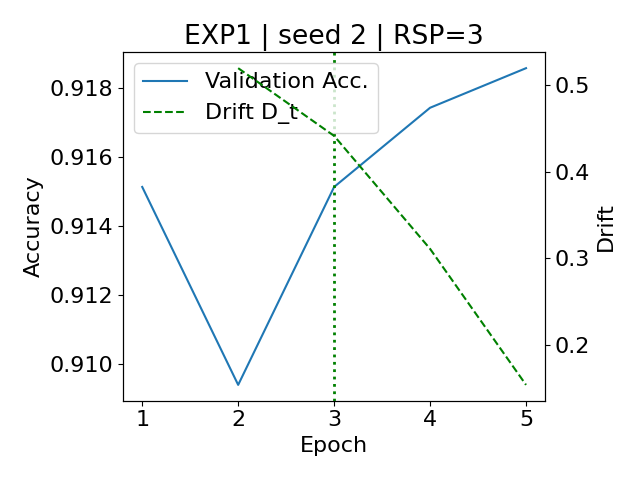}
        \caption{~\textsc{MiniLM} + SST-2; seed=2}
    \end{subfigure}

    \vspace{1em}

    \begin{subfigure}{0.32\textwidth}
        \includegraphics[width=\linewidth]{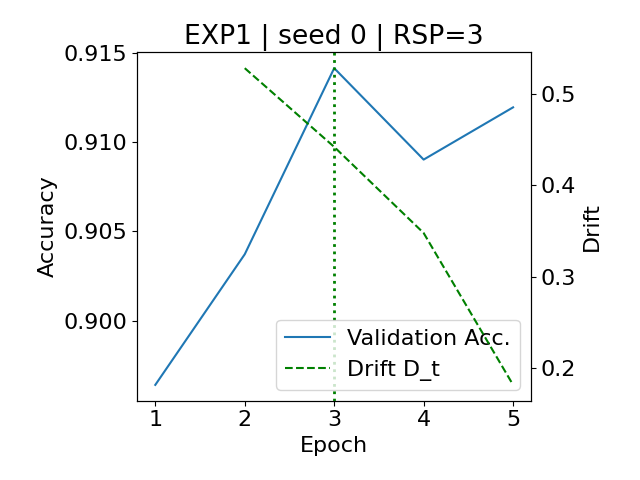}
        \caption{~\textsc{MiniLM} + QNLI; seed=0}
    \end{subfigure}
    \hfill
    \begin{subfigure}{0.32\textwidth}
        \includegraphics[width=\linewidth]{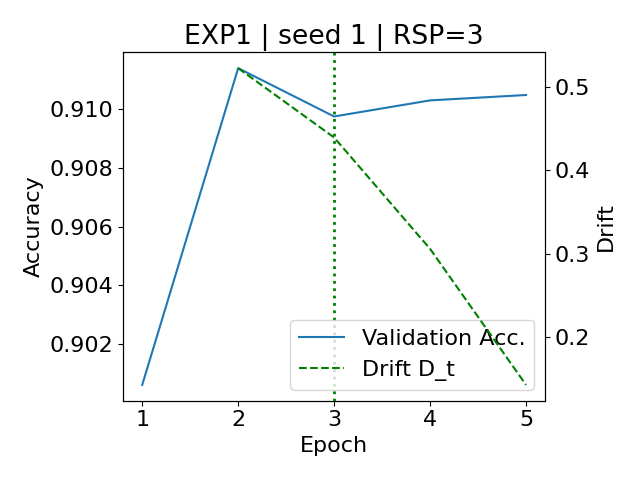}
        \caption{~\textsc{MiniLM} + QNLI; seed=1}
    \end{subfigure}
    \hfill
    \begin{subfigure}{0.32\textwidth}
        \includegraphics[width=\linewidth]{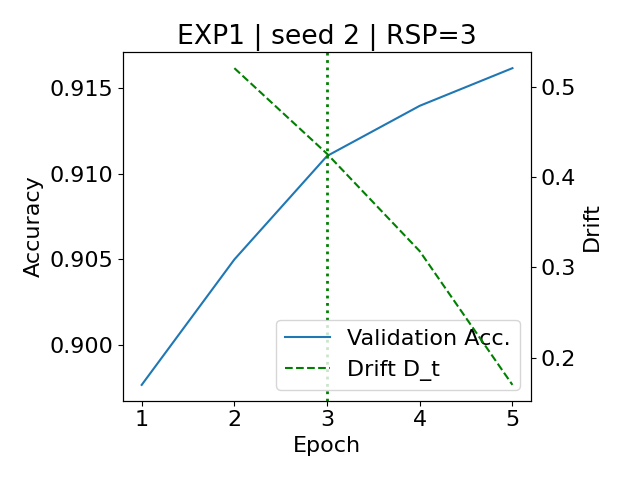}
        \caption{~\textsc{MiniLM} + QNLI; seed=2}
    \end{subfigure}

    \caption{\textbf{Experiment 1 Full Results:} Validation accuracy (solid) and explanation drift $D_t$ (dashed) across the two settings. The dotted vertical line marks the Reasoning Stabilization Point (RSP; Eq.~\ref{eq:rsp}). Accuracy changes only marginally after RSP, consistent with explanations stabilizing by (or before) accuracy saturation.}
    \label{fig:exp1}
\end{figure*}

Figure~\ref{fig:exp1} plots validation accuracy and drift across epochs. We observe a consistent drop in drift early in training, followed by a low-drift regime beginning at RSP (vertical line). Quantitatively, as listed in Table~\ref{tab:main}, RSP occurs at epoch $3.0$ in all our runs, while accuracy changes only slightly after RSP occurrence. This supports our claim that attribution patterns become stable early, and that post-RSP fine-tuning yields diminishing returns in in-domain accuracy despite continued training.

\subsection{Experiment 2: Drift reveals shortcut adoption}
\label{subsec:exp2}
\begin{figure*}[t] 
    \centering
    
    \begin{subfigure}{0.32\textwidth}
        \includegraphics[width=\linewidth]{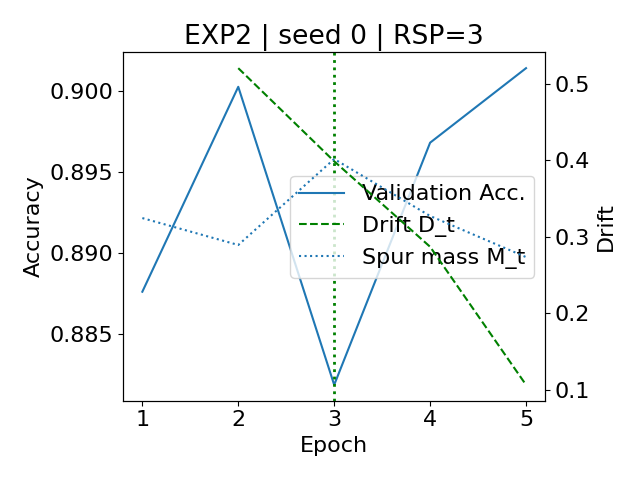}
        \caption{~\textsc{DistilBERT} + SST-2; seed=0}
    \end{subfigure}
    \hfill
    \begin{subfigure}{0.32\textwidth}
        \includegraphics[width=\linewidth]{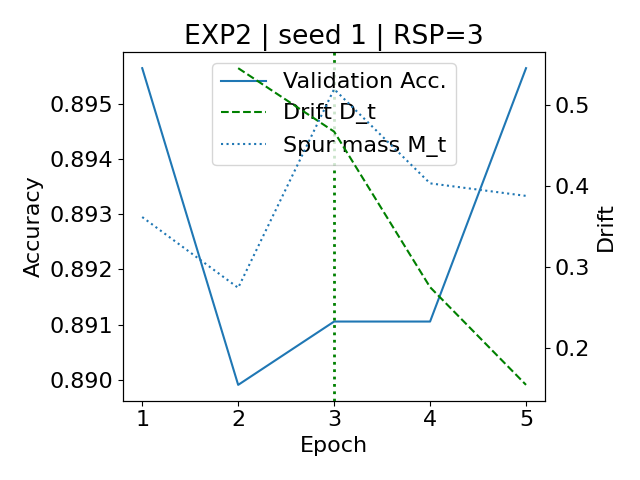}
        \caption{~\textsc{DistilBERT} + SST-2; seed=1}
    \end{subfigure}
    \hfill
    \begin{subfigure}{0.32\textwidth}
        \includegraphics[width=\linewidth]{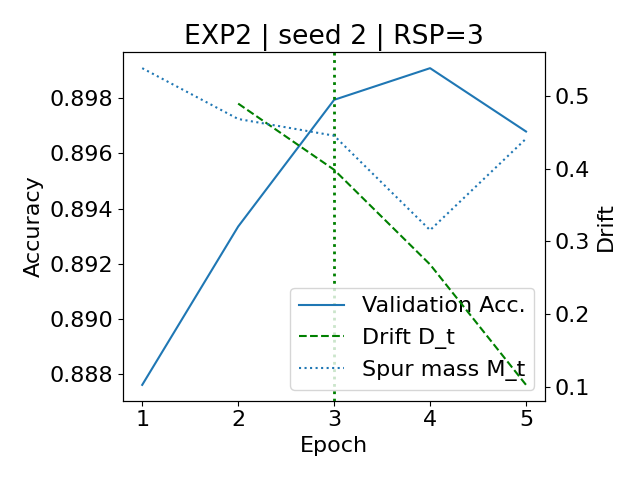}
        \caption{~\textsc{DistilBERT} + SST-2; seed=2}
    \end{subfigure}

    \vspace{1em} 

    \begin{subfigure}{0.32\textwidth}
        \includegraphics[width=\linewidth]{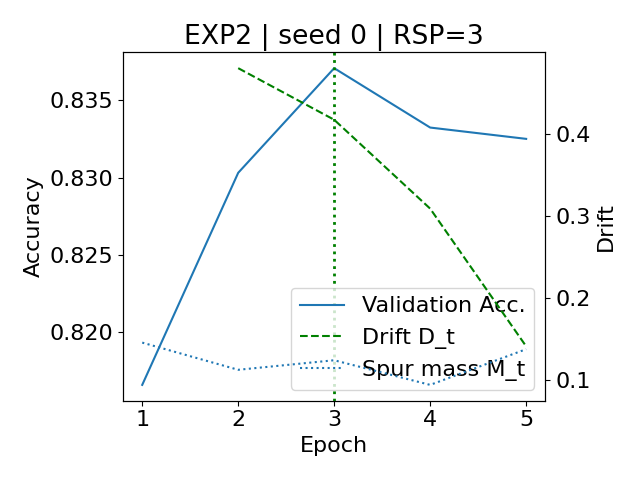}
        \caption{~\textsc{DistilBERT} + QNLI; seed=0}
    \end{subfigure}
    \hfill
    \begin{subfigure}{0.32\textwidth}
        \includegraphics[width=\linewidth]{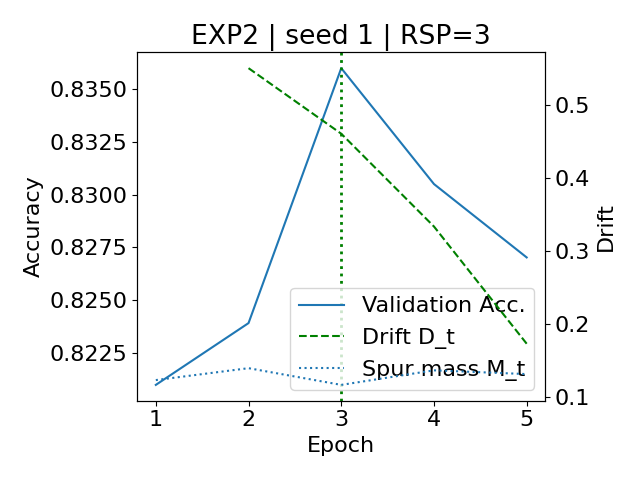}
        \caption{~\textsc{DistilBERT} + QNLI; seed=1}
    \end{subfigure}
    \hfill
    \begin{subfigure}{0.32\textwidth}
        \includegraphics[width=\linewidth]{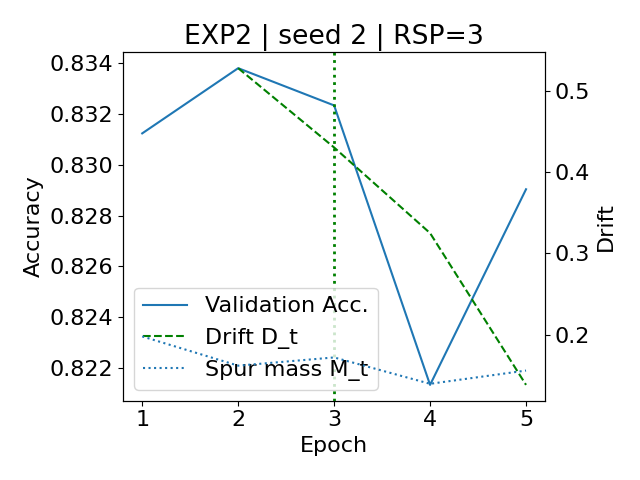}
        \caption{~\textsc{DistilBERT} + QNLI; seed=2}
    \end{subfigure}

    \vspace{1em}

    \begin{subfigure}{0.32\textwidth}
        \includegraphics[width=\linewidth]{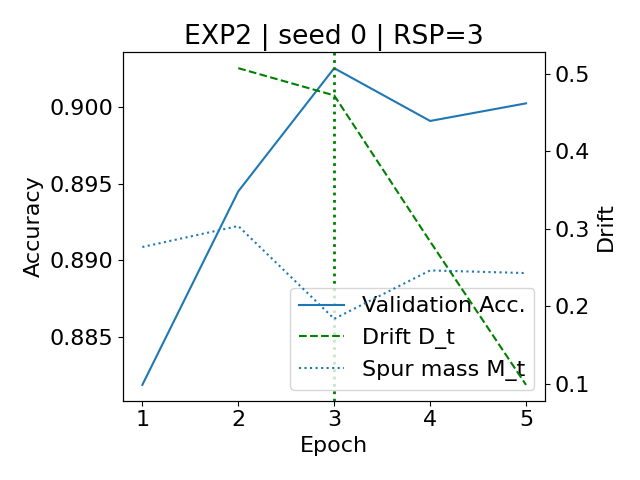}
        \caption{~\textsc{MiniLM} + SST-2; seed=0}
    \end{subfigure}
    \hfill
    \begin{subfigure}{0.32\textwidth}
        \includegraphics[width=\linewidth]{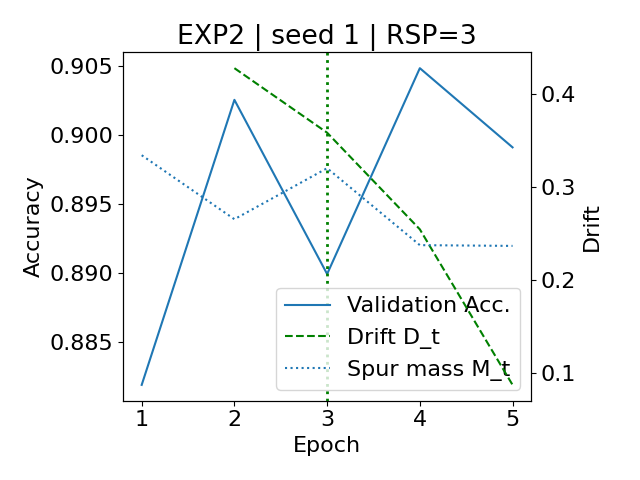}
        \caption{~\textsc{MiniLM} + SST-2; seed=1}
    \end{subfigure}
    \hfill
    \begin{subfigure}{0.32\textwidth}
        \includegraphics[width=\linewidth]{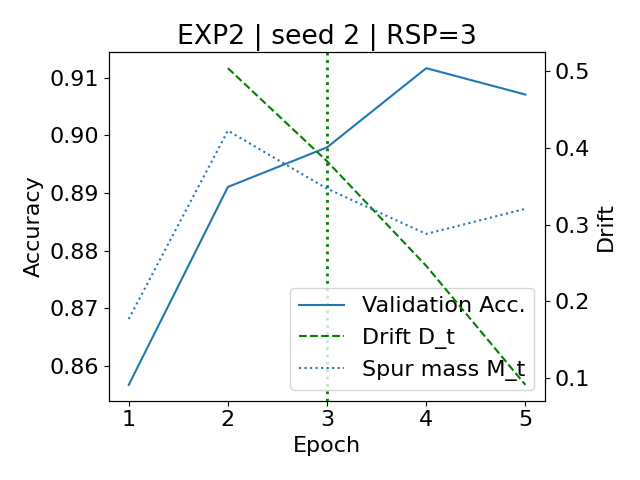}
        \caption{~\textsc{MiniLM} + SST-2; seed=2}
    \end{subfigure}

    \vspace{1em}

    \begin{subfigure}{0.32\textwidth}
        \includegraphics[width=\linewidth]{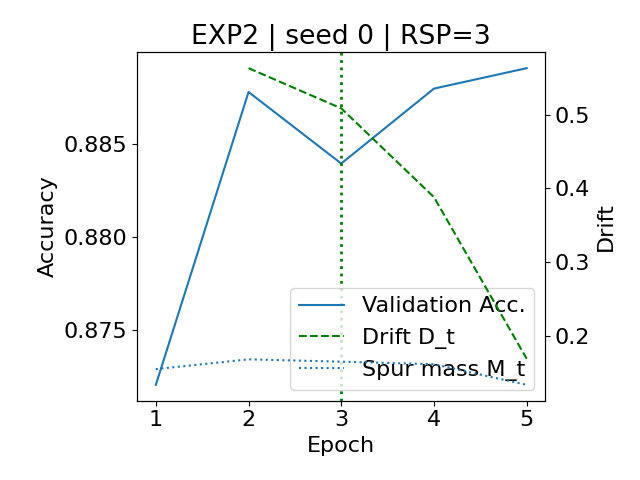}
        \caption{~\textsc{MiniLM} + QNLI; seed=0}
    \end{subfigure}
    \hfill
    \begin{subfigure}{0.32\textwidth}
        \includegraphics[width=\linewidth]{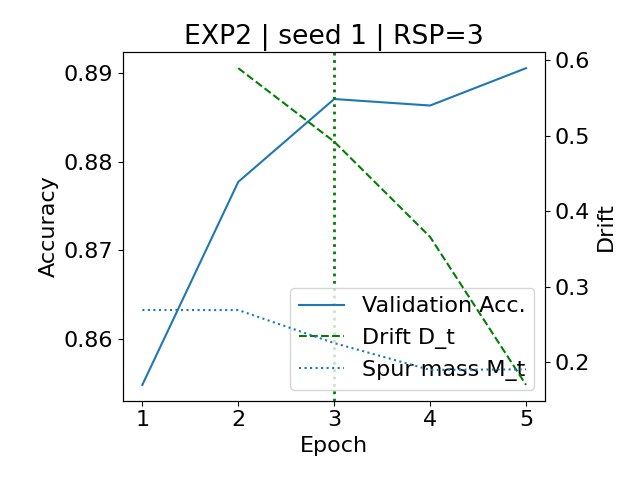}
        \caption{~\textsc{MiniLM} + QNLI; seed=1}
    \end{subfigure}
    \hfill
    \begin{subfigure}{0.32\textwidth}
        \includegraphics[width=\linewidth]{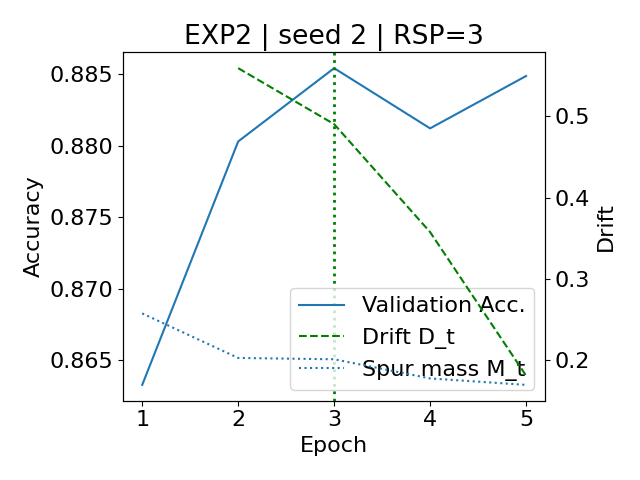}
        \caption{~\textsc{MiniLM} + QNLI; seed=2}
    \end{subfigure}

    \caption{\textbf{Experiment 2 Full Results:} Validation accuracy (solid), drift $D_t$ (dashed), and spur attribution mass $M_t$ (dotted; Eq.~7) across epochs for a representative seed. The spur token is injected into training with label correlation (validation remains clean). $M_t$ increases/peaks near RSP, indicating growing reliance on the shortcut feature despite competitive validation accuracy.}
    \label{fig:exp2}
\end{figure*}

To test whether drift captures shortcut dynamics, we create a controlled spurious correlation in the training set: with probability $p{=}0.8$, we prepend a label-correlated token (\texttt{[SPUR\_POS]} or \texttt{[SPUR\_NEG]}) to each training example. Validation remains \emph{clean} (no spur tokens). In addition to drift, we compute \emph{spur attribution mass} on a small spur-probe set (200 spurious training examples):
\begin{equation}
M_t \;=\; \mathbb{E}_{x\in P_{\mathrm{spur}}}\left[\sum_{j:\,w_j=s(x)} \tilde{E}_{\theta_t}(x)_j\right],
\end{equation}
i.e., the fraction of total attribution assigned to the injected spur token. Figure~\ref{fig:exp2} shows that $M_t$ peaks near RSP while validation accuracy remains high, indicating that explanation dynamics can surface reliance on shortcut features even when standard accuracy metrics appear stable. Similar to Experiment~\ref{subsec:exp1} and as listed in Table~\ref{tab:main}, RSP occurs at epoch $3.0$ in all our runs, while accuracy changes only slightly after RSP occurrence. These results suggest that explanation drift provides an interpretable training-time view of when the model's decision evidence stabilizes and when it reallocates importance toward brittle cues.

\begin{table}[t]
\centering
\small
\setlength{\tabcolsep}{3pt} 
\begin{tabular}{lcccc}
\toprule
Exp. & Setting & RSP & Acc@RSP & Peak Acc \\
\midrule
\multirow{2}{*}{1} & \textsc{DistilBERT} + & \multirow{2}{*}{3.0} & \multirow{2}{*}{90.21 $\pm$ 0.48} & \multirow{2}{*}{91.21 $\pm$ 0.59} \\
                   & SST-2 & & & \\

\multirow{2}{*}{1} & \textsc{DistilBERT} + & \multirow{2}{*}{3.0} & \multirow{2}{*}{88.08 $\pm$ 0.29} & \multirow{2}{*}{88.91 $\pm$ 0.22} \\
                   & QNLI & & & \\

\multirow{2}{*}{1} & \textsc{MiniLM} + & \multirow{2}{*}{3.0} & \multirow{2}{*}{91.40 $\pm$ 0.41} & \multirow{2}{*}{92.09 $\pm$ 0.36} \\
                   & SST-2 & & & \\

\multirow{2}{*}{1} & \textsc{MiniLM} + & \multirow{2}{*}{3.0} & \multirow{2}{*}{91.16 $\pm$ 0.23} & \multirow{2}{*}{91.39 $\pm$ 0.24} \\  
                   & QNLI & & & \\

\midrule
\multirow{2}{*}{2} & \textsc{DistilBERT} + & \multirow{2}{*}{3.0} & \multirow{2}{*}{88.03 $\pm$ 0.81} & \multirow{2}{*}{89.87 $\pm$ 0.29} \\
                   & SST-2 & & & \\

\multirow{2}{*}{2} & \textsc{DistilBERT} + & \multirow{2}{*}{3.0} & \multirow{2}{*}{83.51 $\pm$ 0.25} & \multirow{2}{*}{84.56 $\pm$ 0.17} \\
                   & QNLI & & & \\

\multirow{2}{*}{2} & \textsc{MiniLM} + & \multirow{2}{*}{3.0} & \multirow{2}{*}{89.68 $\pm$ 0.64} & \multirow{2}{*}{90.63 $\pm$ 0.48} \\
                   & SST-2 & & & \\

\multirow{2}{*}{2} & \textsc{MiniLM} + & \multirow{2}{*}{3.0} & \multirow{2}{*}{88.55 $\pm$ 0.16} & \multirow{2}{*}{88.82 $\pm$ 0.29} \\
                   & QNLI & & & \\
\bottomrule
\end{tabular}
\caption{Summary over seeds (mean$\pm$std). See Fig.~\ref{fig:exp1}--\ref{fig:exp2} (provided in Appendix) for drift/accuracy trajectories and RSP.}
\label{tab:main}
\end{table}

\section{Conclusion}
\label{sec:conclusion}
We presented a training-time perspective on interpretability by modeling token-level explanations as a time series during fine-tuning. Our core diagnostic, \emph{explanation drift}, measures epoch-to-epoch changes in attribution patterns on a fixed probe set, while the \emph{Reasoning Stabilization Point (RSP)} summarizes this trajectory as the earliest epoch after which drift remains consistently low. Across multiple datasets and lightweight transformer models, we observed that drift typically stabilizes early, while validation accuracy changes only marginally thereafter.

We further showed that explanation drift can surface shortcut adoption under controlled spurious correlations: attribution mass shifts toward the shortcut feature near RSP even when validation accuracy remains competitive. Together, these results suggest that explanation drift provides a simple, low-cost complement to accuracy curves, offering insight into \emph{when} a model’s decision evidence stabilizes and \emph{how} that evidence reallocates during fine-tuning. We view stabilization signals such as RSP as a practical tool for monitoring training dynamics and guiding checkpoint selection in fine-tuned NLP models.

\section{Limitations}
\label{sec:limitations}
Our study focuses on lightweight transformer models and small-to-medium classification benchmarks, with fine-tuning runs limited to a small number of epochs. While this setting is sufficient to reveal clear explanation drift dynamics, the behavior of drift and the Reasoning Stabilization Point (RSP) in larger models, longer training regimes, or generative tasks remains an open question. In particular, models trained for many epochs or with curriculum-style schedules may exhibit multiple stabilization phases that are not captured by a single RSP.

Explanation drift depends on the choice of explanation function and similarity metric. Although we use standard gradient-based attributions and rank-based similarity for robustness, different attribution methods may yield quantitatively different drift curves. Our results, therefore, characterize stabilization with respect to a given explainer, rather than providing an explainer-independent guarantee. In addition, our current analysis focuses on token-level evidence and does not directly address higher-level linguistic structures or compositional reasoning.

Finally, while controlled shortcut injection offers a clear diagnostic setting, real-world spurious correlations may be subtler and harder to isolate. Explanation drift should therefore be viewed as a complementary signal rather than a standalone criterion for robustness or correctness.

\section{Ethical considerations}
\label{sec:ethical_considerations}
This work introduces a diagnostic method for analyzing training dynamics and explanation stability in neural language models. It does not propose or evaluate a deployed decision-making system, does not involve interaction with users or human subjects, and does not introduce new risks related to data privacy, user interaction, or downstream automation. All experiments are conducted on standard, publicly available benchmark datasets and synthetic modifications that do not contain sensitive or personal information.

As such, we do not identify any new ethical risks arising from this work.

\section{Notes From Authors}
\label{sec:notes_from_authors}
The authors used AI Assistants for this research. Particularly, ChatGPT was used for coding and grammar, and Copilot was used for automatic completion during coding.

\bibliography{custom}


\appendix




\section{Appendix: Plots for Experiments 1 and 2}
\label{sec:appendix_plots_for_experiment_1}

\end{document}